\documentclass[10pt,twocolumn,letterpaper]{article}

\usepackage[pagenumbers]{cvpr} 

\usepackage{graphicx}
\usepackage{amsmath}
\usepackage{amssymb}
\usepackage{booktabs}

\usepackage{bm}
\usepackage{bbm}

\usepackage{algorithm}
\usepackage{algorithmic}
\usepackage[pagebackref,breaklinks,colorlinks]{hyperref}

\usepackage[capitalize]{cleveref}
\crefname{section}{Sec.}{Secs.}
\Crefname{section}{Section}{Sections}
\Crefname{table}{Table}{Tables}
\crefname{table}{Tab.}{Tabs.}

\def\cvprPaperID{6548} 
\def\confName{CVPR}
\def\confYear{2023}

\begin{document}

\title{Few-shot Non-line-of-sight Imaging\\
with Signal-surface Collaborative Regularization}

\author{Xintong Liu$^{1}$, Jianyu Wang$^{1}$, Leping Xiao$^{1}$, Xing Fu$^{1}$, Lingyun Qiu$^{1,2}$, Zuoqiang Shi$^{1,2}$\\
\\
1 Tsinghua University\\
2 Yanqi Lake Beijing Institute of Mathematical Sciences and Applications
}
\maketitle

\begin{abstract}
    The non-line-of-sight imaging technique aims to reconstruct targets from multiply reflected light. For most existing methods, dense points on the relay surface are raster scanned to obtain high-quality reconstructions, which requires a long acquisition time. In this work, we propose a signal-surface collaborative regularization (SSCR) framework that provides noise-robust reconstructions with a minimal number of measurements. Using Bayesian inference, we design joint regularizations of the estimated signal, the 3D voxel-based representation of the objects, and the 2D surface-based description of the targets. To our best knowledge, this is the first work that combines regularizations in mixed dimensions for hidden targets. Experiments on synthetic and experimental datasets illustrated the efficiency and robustness of the proposed method under both confocal and non-confocal settings. We report the reconstruction of the hidden targets with complex geometric structures with only $5 \times 5$ confocal measurements from public datasets, indicating an acceleration of the conventional measurement process by a factor of 10000. Besides, the proposed method enjoys low time and memory complexities with sparse measurements. Our approach has great potential in real-time non-line-of-sight imaging applications such as rescue operations and autonomous driving.
\end{abstract}

\section{Introduction}
\label{sec:intro}
\begin{figure}[t]
  \centering
   \includegraphics[width=1\linewidth]{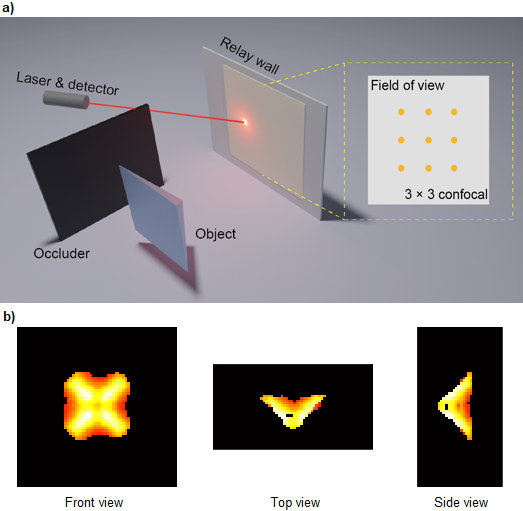}
   \caption{A typical non-line-of-sight imaging scenario. a) The time resolved signals are measured at only $3 \times 3$ focal points. b) The three views of the reconstructed target obtained with the proposed signal-surface collaborative regularization (SSCR) method.}
   \label{fig:layout}
\end{figure}

The non-line-of-sight (NLOS) imaging technique enables reconstructions of targets out of the direct line of sight, which is attractive in various applications such as autonomous driving, remote sensing, rescue operations and medical imaging \cite{Street,Polarized,Fermat,Tracking,Pose,Perspective,Long,Picosecond,Acoustic,Thermal_spe_dif,Thermal,Loc_iden,2D,Photon-efficient}. A typical scenario of NLOS imaging is shown in \Cref{fig:layout}. Several points on the visible surface are illuminated by a laser and the back-scattered light from the target is detected to reconstruct the target. The time-correlated single-photon counting (TCSPC) technique is applied in the detection process due to the extremely low photon intensity after multiple diffuse reflections. In practice, a single-photon avalanche diode (SPAD) in the Geiger-mode can be used to record the photon events with time-of-flight (TOF) information \cite{SPAD}. The first experimental demonstration of NLOS imaging dates back to 2012, where the targets are reconstructed with the back-projection (BP) method \cite{First_work_NLOS}. Extensions of this approach include its fast implementation \cite{FBP}, the filtering technique for reconstruction quality enhancement \cite{LOG_BP}, and weighting factors for noise reduction \cite{Weight}. The BP-type methods have high computational complexity, resulting in slow reconstruction processes. 
A number of efficient methods have been designed for fast reconstructions. The light cone transform (LCT) method \cite{LCT} formulates the physical model as a convolution operator, so that the reconstructions can be obtained using the Wiener deconvolution method with fast Fourier transform. The directional light cone transform (D-LCT) \cite{DLCT} generalizes the LCT and reconstructs the albedo and surface normal simultaneously. The method of frequency wavenumber migration (F-K) \cite{FK} formulates the propagation of light using the wave equation, and also provides a fast inversion algorithm with the frequency-domain interpolation technique. Whereas the LCT, D-LCT and F-K methods only work directly in confocal measurement scenarios, the phasor field (PF) method \cite{Phasor_wave,Virtual,Phasor} converts the NLOS imaging scenarios to LOS cases and works for the general non-confocal setting with low computation complexity. For high-quality and noise-robust reconstructions, the signal-object collaborative regularization (SOCR) method can be applied, but brings additional computational cost. In recent years, deep learning-based methods are also introduced to the field of NLOS imaging \cite{Steady_state,Deep,Deep_inverse,Deep_remapping}. Besides, advances in hardware enhance the distance of NLOS detection to kilometers \cite{Long}, or make it possible to reconstruct targets on the scale of millimeters\cite{Picosecond}.

Despite these breakthroughs, the trade off between the acquisition time and the imaging quality is inevitable. In the raster scanning mode, the acquisition time is proportional to the number of measurement points and the number of pulses used for each illumination. Due to the intrinsic ill-posedness of the NLOS reconstruction problem\cite{Analysis} and heavy measurement noise\cite{Weight}, dense measurements are necessary for high quality reconstructions\cite{LCT,FK,Phasor}. The measurement process may take from seconds to hours, which poses a great challenge for applications such as autonomous driving, where real-time reconstruction of the video stream is needed. The acquisition process can be accelerated by reducing the number of pulses used for each illumination point. In the work \cite{FF}, the pulse number that record the first returning photon is used to reconstruct the target. Another way to reduce the acquisition time is to design array detectors for non-confocal measurements. For example, the implementation of the phasor field method with SPAD arrays realizes low-latency real-time video imaging of the hidden scenes \cite{LowLatency}. A third way to accelerate the NLOS detection process is to reduce the number of measurement pairs. It is shown that $16 \times 16$ confocal measurements are enough to reconstruct the hidden target by incorporating the compressed sensing technique \cite{Compressed}.


In this paper, we study the randomness in the photon detection process of non-line-of-sight scenarios and propose an imaging method that deals with a very limited number of measurements. We design joint regularizations of the estimated signal, the 3D voxel-based representation of the objects, and the 2D surface-based description of the targets, which leads to faithful reconstruction results. The main contributions of our work are as follows.
\begin{itemize}
    \item We propose a novel signal-surface collaborative regularization (SSCR) framework for few-shot non-line-of-sight reconstructions, which works under both confocal and non-confocal settings.
    \item We report the reconstruction of the hidden targets with complex geometric structures with only $5 \times 5$ confocal measurements from public datasets, indicating an acceleration of the conventional measurement process by a factor of 10000.
\end{itemize}

\begin{figure}[t]
    \centering
     \includegraphics[width=1\linewidth]{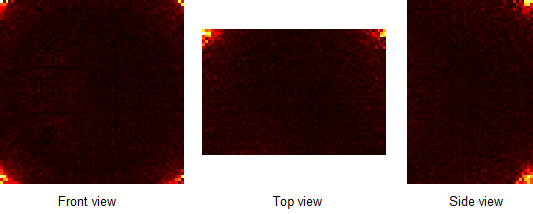}
     \caption{The least squares solution of the statue with $3 \times 3$ confocal measurements\cite{FK}. The solution does not contain useful information of the target even though its simulated signal matches the measurements well (see \cref{fig2:L2_signal}). Strong regularizations are needed to reconstruct the target. See also \cref{fig6:multirec} for a comparison.}
     \label{fig3:L2_solution}
  \end{figure}

\begin{figure}[t]
    \centering
     \includegraphics[width=1\linewidth]{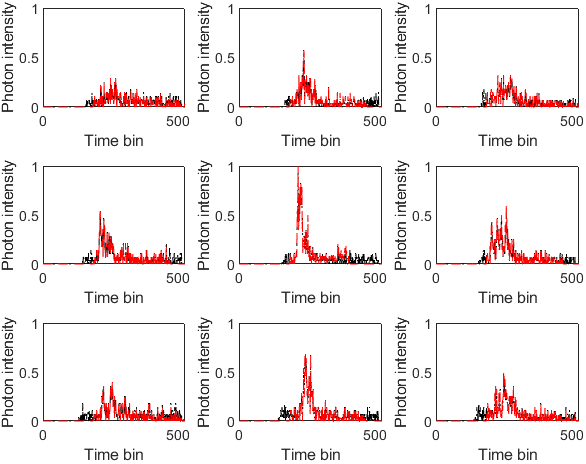}
     \caption{Comparisions of the measured data and the simulated data of the least squares solution for the instance of the statue\cite{FK}. The measured signals are shown in black. The simulated data of the least squares solution are shown in red. The shapes of the signals are very close to each other.}
     \label{fig2:L2_signal}
  \end{figure}

\section{Related work}
From a mathematical point of view, the non-line-of-sight imaging task belongs to the category of inverse problem. The goal is to reconstruct the surface of the hidden target with the measured signal. The NLOS inverse problem is ill-posed due to the intrinsic structure of the physical model\cite{Analysis} and heavy measurement noise\cite{Weight}. When the number of measurement points is small, the lack of data leads to rank-deficiency of the measurement matrix, making the reconstruction task even harder\cite{Compressed}. In such cases, regularization becomes a vital tool for high-quality reconstructions. In \cref{fig3:L2_solution} and \cref{fig2:L2_signal} we show the least squares solution without regularization for the instance of the statue with $3 \times 3$ confocal measurements \cite{FK} and compare the simulated signal with the raw measurements. The simulated signal matches the input data well, but the target cannot be identified in the reconstruction. More details are provided in the supplement. Here, we review two regularization based methods that provide high-quality reconstructions. The proposed method is closely related with these methods.

\vspace{0.15cm}

\noindent\textbf{The SOCR method}\hspace{1em}The signal-object collaborative regularization (SOCR) method \cite{SOCR} represents the hidden target with voxels and considers prior terms for both the reconstructed target and the signal. The joint regularization term concerns sparseness of the reconstruction, non-local self-similarity of the target and the smoothness of the signal. The method employs $L_1$ regularization, orthogonal optimization and empirical Wiener filter to provide noiseless reconstructions with sharp boundaries, but at the expense of additional computational cost.

\vspace{0.15cm}

\noindent\textbf{The CC-SOCR method}\hspace{1em}The SOCR method only works in cases where the signals are measured at rectangular grid points. For generalizations to cases with arbitrary illumination and detection pattern, the confocal-complemented signal-object collaborative regularazation (CC-SOCR) method \cite{CC-SOCR} considers the virtual confocal signal measured at a focal plane to overcome the rank-deficiency. Joint sparse representation of the simulated signal and the virtual confocal signal is used to increase robustness. It was reported that the method can reconstruct the targets with 200 confocal measurements.

Despite these considerations, reconstruction of the hidden targets may fail when only a very limited number of measurements (for example, $5 \times 5$ confocal measurements) are available. In such cases, the ill-posedness of the NLOS inverse problem is so strong that these methods provide biased estimations of the targets. We propose a novel signal-surface collaborative regularization (SSCR) method that provides faithful reconstructions under few-shot scenarios. The proposed approach are closely related with SOCR and CC-SOCR methods but improves these methods in the following two aspects.
\begin{itemize}
    \item \textbf{The data fidelity term} These methods use the quadratic data misfit as the data fidelity terms, which assume noise of the Gaussian type. In this work, we study the randomness of photon event stamping, and use logarithmic data fidelity terms derived from Bernoulli photon event assumptions.
     
    \item \textbf{The dimensionality of priors} These methods represent the hidden targets with three-dimensional voxels and the regularizations apply to three-dimensional tensors. In fact, we aim at reconstructing the surface of the hidden target, which is a two-dimensioanl geometric object in nature. In this work, we study the intrinsic two-dimensional regularization for the hidden surface and incorporate priors in both 2D and 3D representations of the targets.
\end{itemize}

\section{The physical model}
In general, the reconstruction quality and the computation complexity of an imaging algorithm rely heavily on the forward model that simulates the physical measurement process. Fine physical models lead to reconstructions with clear geometric structures, but at the expense of high computational cost\cite{Occluder,Beyond}. Instead of putting forward a novel physical model, we aim at overcoming the ill-posedness of the NLOS imaging task with an extremely small number of measurements. We adopt a linear physical model which only considers the square fall-off of the photon intensity. Let $x'_i$ and $x'_d$ be the illumination and detection points on the visible surface, the photon intensity detected at time $t$ is modeled as
\begin{equation}
\label{general_physical_model}
\begin{aligned}
\tau(x'_i,x'_d,t) =&\int_{\Omega}{\frac{u(x)}{\| x'_{i}-x \| ^2 \| x'_{d}-x\|^2}}\cdot\\
&\ \ \ \ \ \ \ \ \ \ \delta (\| x'_{i}-x\| +\left\| x'_{d}-x \right\| -ct ) dx,
\end{aligned}
\end{equation}
in which $c$ is the speed of light, $x$ is a point in the three-dimensional reconstruction domain $\Omega$. The albedo value of the point $x$ is represented by $u(x)$. The $\delta$ function describes the intrinsic domain of integration as the set of points with optical path length $ct$. When $x'_i \neq x'_d$, the domain of integration is a half ellipsoid with foci $x'_i$ and $x'_d$. When $x'_i = x'_d = x'$, the domain of integration is a half sphere with center $x'$ and the model reduces to the one used in the work \cite{LCT} given below
\begin{equation}
    \label{confocal_physical_model}
    \tau(x',t) =\int_{\Omega}{\frac{u(x)}{\| x'-x \| ^4}}\cdot\delta (2\| x'-x\|- ct ) dx.
\end{equation}

\section{The proposed framework}
In this section, we study the randomness in the measurement process and propose a novel signal-surface collaborative regularization (SSCR) framework for few-shot NLOS imaging.

\vspace{0.15cm}

\noindent\textbf{Notation}\hspace{1em}The $L_2$ norm of a vector $\mathbf{x}$ is denoted by $\|\mathbf{x}\|$. We use $[N]$ as an abbreviation of the set $\{1,2,\dots,N\}$. The three-dimensional reconstruction domain $\Omega$ is discretized with voxels $V = \{v_{ijk}=(x_i,y_j,z_k)|i\in[I],j\in[J],k\in[K]\}$, where $x_i$, $y_j$, and $z_k$ are the coordinates of the point $v_{ijk}$ in the horizontal, vertical and depth directions. Each point in $V$ represents a cubic voxel of the same size. The grid function $\mathbf{u}$ is used to denote the discrete albedo values, with its components denoted by $u_{ijk}=\mathbf{u}(v_{ijk})$. To reconstruct the hidden targets, the signal is measured at $P$ pairs of points $\{(x'_p,y'_p)\}^P_{p=1}$, in which $x'_p$ and $y'_p$ are the coordinates of the $p^{th}$ illumination and detection points, respectively. For each measurement pair $(x'_p,y'_p)$, the one-dimensional time resolved signal contains $Q$ time bins. The length of each time bin is a constant, usually at the scale of picoseconds in real applications. We denote by $\tau_{p,q}$ the photon intensity detected at the $p^{th}$ measurement pair and the $q^{th}$ time bin. The measured dataset is denoted by $\bm{\tau}$.

\vspace{0.15cm}

\noindent\textbf{The reconstructed surface}\hspace{1em}We use $\mathbf{u}\in \mathbb{R}^{I \times J \times K}$ to represent the albedo of the NLOS scene, which is a three-diminsional tensor. However, it is only possible to reconstruct the portion of the hidden surface where photons are bounced back, which is a two dimenional geometric object. With this observation, we define a subset $\mathcal{G}\subsetneq \mathbb{R}^{I \times J \times K}$ as follows
\begin{equation}
\label{G}
\begin{aligned}
\mathcal{G}=&\{\mathbf{g}=(g_{ijk})\in \mathbb{R}^{I \times J \times K}\ |\ \forall (i,j)\in[I]\times[J],\\
&\ \ \ \ \ \ \exists \text{ at most one } k = k_{ij}\in[K], \text{ s.t. }g_{i,j,k_{ij}} \neq 0\}.
\end{aligned}
\end{equation}
Each element $\mathbf{g}\in\mathcal{G}$ yields a trival two-dimensional parameterization. For each pixel $(i,j)\in [I] \times [J]$, only one of the following cases holds.
\begin{itemize}
    \item Case 1: $g_{ijk} = 0$ for all $k\in[K]$. In this case, the line $x = x_i, y = y_j$ does not intersect with the target and we call $(i,j)$ a \emph{background pixel}.
    \item Case 2: There exists only one $k = k_{ij}$, such that $g_{i,j,k_{ij}}>0$. In this case, we call $(i,j)$ a \emph{foreground pixel} and the corresponding $depth$ is $z_{ij}$.
\end{itemize}
To express elements of $\mathcal{G}$ with matrices, we show that there is a bijection from the set $\mathcal{G}$ to the following set
\begin{equation}
    \label{G'}
    \begin{aligned}
    \mathcal{G}'=\{(\mathbf{e}, \mathbf{d}, \bm{\alpha})\ |\ \mathbf{e} &= (e_{ij})_{I \times J}, e_{ij} \in \{0,1\},\\ 
    \mathbf{d}&=(d_{ij})_{I \times J}, d_{ij} \in [K] \cup \{\operatorname{NaN}\},\\
    \bm{\alpha}&=(\alpha_{ij})_{I \times J}, \alpha_{ij} \in{\mathbb{R}} \cup \{\operatorname{NaN}\},\\
    \forall (i,j),\ \ e_{ij} = 0 &\iff d_{ij} = \operatorname{NaN}\iff \alpha_{ij} = \operatorname{NaN}\}.
    \end{aligned}
    \end{equation}
In the definition of $\mathcal{G}'$, the placeholder \text{NaN} represents a background pixel and does not operate with real numbers. To construct a bijection from $\mathcal{G}$ to $\mathcal{G}'$, for each element $g \in \mathcal{G}$, let $\mathbf{e}$ be the indicator function of the set of foreground pixels, $\mathbf{d}$ be the depths of the foreground pixels and $\bm{\alpha}$ be the corresponding albedo values. Then, fill the matrices $\mathbf{d}$ and $\bm{\alpha}$ with \text{NaN}s where necessary. It is easy to check that this map: $\mathcal{G}\rightarrow \mathcal{G}'$ is one to one and onto. We call $\mathbf{e}$, $\mathbf{d}$ and $\bm{\alpha}$ the indicator matrix, the depth matrix and the albedo matrix of $\mathbf{g}$, respectively. For simplicity, we also denote $\mathcal{G}'$ as $\mathcal{G}$. The set $\mathcal{G}$ is still too large to capture the set of real-world surfaces, as the distribution of the foreground pixels and depths can be disorganized. In the following sections, we will design regularizations for elements of $\mathcal{G}$, where the matrix representations bring remarkable convenience.

\vspace{0.15cm}

\noindent\textbf{The photon event histogram}\hspace{1em}In NLOS scenarios, the intensity of the back scattered light is extremely weak after multiple diffuse reflections. For each measurement pair, a total of $N$ laser pulses are emitted to the illumination point, and the TCSPC device can be used to record the photon events. Commonly used detectors are single photon avalanche diodes (SPAD). Although heavily corrupted with background noise, the photon event histogram provides an estimation of the photon intensity. We use the binary variables $d_{p,q,n}$ to denote the recorded photon events. 
\begin{equation}
    d_{p,q,n}=\left\{ \begin{array}{l}
	1,\ \mathrm{record}\ \mathrm{a}\ \mathrm{photon}\ \mathrm{event}\\
	0,\ \mathrm{otherwise}\\
\end{array} \right., 
\end{equation}
in which $p\in[P]$, $q\in[Q]$, and $n\in[N]$ are indices of the measurement pair, the time bin and the pulse number. For each measurement pair, the detector can record at most one photon event in each pulse, which means that $\sum_q d_{p,q,n} \leq 1$. However, the cases of recording more than one photon event in a single pulse can be neglected \cite{FF}. The collection of photon event stamping is denoted by $\mathbf{d}$.

\vspace{0.15cm}

\noindent\textbf{The Bayesian framework}\hspace{1em}We propose a unified Bayesian framework that reconstructs the hidden surface with the data of photon event stamping. For each measurement pair $p$, time bin $q$, and pulse number $n$, it is assumed that the detection of a photon event $e_{p,q,n}$ follows the Bernoulli distribution with probability $\mathbb{P} \left\{ e_{p,q,n}=1 \right\} = 1-e^{-\eta \tau _{p,q}}$, in which $\eta > 0$ is the detection efficiency \cite{LCT}. The collection of random variables $e_{p,q,n}$ is denoted by $\mathbf{e}$. In NLOS detection scenarios, the probability of detecting a photon event is extremely small. The first order approximation of the exponent is adopted and we assume 
\begin{equation}
\label{e_tau}
    \mathbb{P} \left\{ e_{p,q,n}=1 \right\} = \eta \tau _{p,q}.
\end{equation}
In \cref{general_physical_model}, the photon intensity is linear with respect to the albedo, we choose $\eta = 1$ without loss of generality. We also assume that the detections of different photon events are independent. Let $\mathbf{g}\in \mathcal{G}$ be the reconstructed target. We view $\mathbf{g}$ and $\bm{\tau}$ as random vectors and find them simultaneously by maximizing the posterior probability $\mathbb{P} (\mathbf{g},\bm{\tau}|\mathbf{e} = \mathbf{d})$, where $\mathbf{e}$ is related with $\bm{\tau}$ by \cref{e_tau}. Noting that the set $\mathcal{G}$ is not a convex subset of $\mathbb{R}^{I \times J \times K}$, the resulting optimization problem is non-convex and hard to solve. Besides, the forward model we adopt is an approximation of the ideal one, and contains inevitable bias. To tackle these problems, we introduce the random vector $\mathbf{u}\in \mathbb{R}^{I \times J \times K}$ as an approximation of the surface $\mathbf{g}$ and maximize $\mathbb{P} (\mathbf{g},\mathbf{u},\bm{\tau}|\mathbf{e} = \mathbf{d})$. Using the Bayesian formula, we obtain
\begin{equation}
    \label{MAP}
    \begin{aligned}
    	&\mathop {\arg\max}_{\mathbf{g},\mathbf{u},\bm{\tau}}\mathbb{P}(\mathbf{g}, \mathbf{u}, \bm{\tau}|\mathbf{e} = \mathbf{d})\\
        =&\mathop {\arg\max}_{\mathbf{g},\mathbf{u},\bm{\tau}}\mathbb{P}(\mathbf{e} = \mathbf{d}|\mathbf{g},\mathbf{u}, \bm{\tau})\mathbb{P}(\mathbf{g}, \mathbf{u}, \bm{\tau})\\
        =&\mathop {\arg\max}_{\mathbf{g},\mathbf{u},\bm{\tau}}\mathbb{P}(\mathbf{e} = \mathbf{d}|\bm{\tau})\mathbb{P}(\mathbf{g},\mathbf{u}, \bm{\tau})\\
        =&\mathop {\arg\max}_{\mathbf{g},\mathbf{u},\bm{\tau}}\prod_{p,q,n}{\mathbb{P}(\mathbf{e}_{p,q,n}=\mathbf{d}_{p,q,n}|\tau _{p,q})}\mathbb{P}(\mathbf{g},\mathbf{u}, \bm{\tau})\\
        =&\mathop {\mathrm{arg}\max}_{\mathbf{g},\mathbf{u},\bm{\tau}}\prod_{p,q}{( \tau _{p,q} ) ^{d_{p,q}}( 1-\tau _{p,q} ) ^{N-d_{p,q}}}\mathbb{P}(\mathbf{g},\mathbf{u},\bm{\tau})\\
        =&\mathop {\mathrm{arg}\min}_{\mathbf{g},\mathbf{u},\bm{\tau}}\ \ \sum_{p,q}[{( d_{p,q}-N ) \ln ( 1-\tau _{p,q} )} \\
    	&\ \ \ \ \ \ \ \ \ \ \ \ \ \ \ \ \ \ \ \ \ \ -d_{p,q}\ln ( \tau _{p,q} )] +\Gamma (\mathbf{g}, \mathbf{u}, \bm{\tau}),
    \end{aligned}
\end{equation}
where $d_{p,q} = \sum_{n=1}^N d_{p,q,n}$ is the data of photon event histogram. In the second equality we assume that the conditional probability of $\mathbf{e}$ only depends on $\bm{\tau}$. $\Gamma (\mathbf{g}, \mathbf{u}, \bm{\tau})$ is the joint prior term of $\mathbf{g}$, $\mathbf{u}$ and $\bm{\tau}$.

\section{The joint regularization term}
The joint regularization term $\Gamma (\mathbf{g}, \mathbf{u}, \bm{\tau})$ plays a crucial role in the process of reconstruction. An ingenious design of this term not only results in faithful reconstructions, but also admits low-cost algorithms to solve the optimization problem (\ref{MAP}). In this paper, we assume
\begin{equation}
    \begin{aligned}
    \Gamma (\mathbf{g}, \mathbf{u}, \bm{\tau}) =\ & \lambda\|\bm{\tau} - A\mathbf{u}\|^2\\
     +& J_1(\mathbf{u}) + J_2(\mathbf{u},\bm{\tau}) + J_3(\mathbf{u},\mathbf{g}),
    \end{aligned}
\end{equation}
in which $A$ is the forward operator defined by \cref{general_physical_model}. $J_1$ is the prior of $\mathbf{u}$, $J_2$ is the joint prior of  $\mathbf{u}$ and $\bm{\tau}$, $J_3$ is the joint prior of $\mathbf{u}$ and $\mathbf{g}$. $\lambda$ is a fixed parameter.

\vspace{0.15cm}

\noindent\textbf{The prior $J_1(\mathbf{u})$}\hspace{1em}The priors of the three-dimensional representations of the targets have been widely used in existing works\cite{FK,LCT,sinogram}. Two efficient priors are the sparseness and non-local self-similarity of the objects. In the work \cite{SOCR}, these two priors are considered for the 4D tensor of the directional-albedo. Here, we simplify the approach and directly use $L_1$ norm of the albedo to impose the sparseness of the target. For non-local self-similarity prior, we directly follow the block-matching and sparse representation method in the work \cite{SOCR}. We set
\begin{equation}
    \begin{aligned}
    J_1(\mathbf{u}) =\ & s_u\|\mathbf{u}\|_1\\
     +& \lambda_u\sum_i\left[\|\mathcal{B}\mathbf{u}_i-D_sC_iD_n^T\|^2 + \lambda_{pu}|C_i|_0\right],
    \end{aligned}
\end{equation}  
in which $s_u$, $\lambda_u$, and $\lambda_{pu}$ are fixed parameters. $|\cdot|_0$ denotes the number of nonzero elements. The summation is made over all possible local 3D sub-tensors indexed by $i$. The matrix $\mathcal{B}\mathbf{u}_i$ is constructed by putting the vectorizations of the $i^{th}$ sub-tensor $\mathbf{u}_i$ and its neighbors column by column. The orthogonal matrices $D_s$ and $D_n$ capture the local structure and non-local self-similarity of the albedo $\mathbf{u}$. $C_i$ contains the transform-domain coefficients of the $i^{th}$ block, whose sparseness is imposed by the term $|C_i|_0$. For more details, we refer the readers to \cite{BM3D,DDTF,SOCR}.

\vspace{0.15cm}

\noindent\textbf{The prior $J_2(\mathbf{u},\bm{\tau})$}\hspace{1em}We seek for a joint local sparse representation scheme for the estimated signal $\bm{\tau}$ and simulated signal $A\mathbf{u}$. It is assumed that $\bm{\tau}$ is a three dimensional tensor of size $N_x \times N_y \times Q$, in which $N_x$ and $N_y$ are the number of measurement points in the horizontal and vertical directions. $Q$ is the number of time bins. We call a three-dimensional sub-tensor of $\bm{\tau}$ a local patch. Consider the set of all possible patches of size $r_x \times r_y \times r_q$. We generate the patch dataset $\mathcal{P}(\bm{\tau})$ by generating the vectorization of each patch and put them together column by column. In the work \cite{CC-SOCR}, the patch dataset of the complete virtual confocal signal is sparsely represented by an adaptively learned dictionary. However, when $N_x$ and $N_y$ are small, the number of patches are not sufficient for effective dictionary learning. Here, we use the orthogonal dictionary $D = D_q \otimes Dy \otimes D_x$ as the transform basis, where $D_q$, $D_y$ and $D_x$ are matrices of the discrete consine transform of orders $q$, $y$ and $x$, respectively. The joint prior of $\mathbf{u}$ and $\bm{\tau}$ is given by
\begin{equation}
    \begin{aligned}
    J_2(\mathbf{u},\bm{\tau}) =\ & \lambda_t \|\mathcal{P}(\bm{\tau}) - DS\|^2\\
     +&\lambda_{ut}\|\mathcal{P}(A\mathbf{u})- DS\|^2 + \lambda_{pt}|S|_0,
    \end{aligned}
\end{equation}
in which $\lambda_t$, $\lambda_{ut}$ and $\lambda_{pt}$ are fixed parameters. $\mathcal{P}(A\mathbf{u})$ represents the patch dataset generated by the simulated signal of $\mathbf{u}$. $S$ contains the public transform domain coefficients of $\mathcal{P}(\bm{\tau})$ and $\mathcal{P}(A\mathbf{u})$, whose sparseness is imposed by the $L_0$ term.

\vspace{0.15cm} 

\noindent\textbf{The prior $J_3(\mathbf{u},\mathbf{g})$}\hspace{1em}We express the joint regularization of $\mathbf{u}$ and $\mathbf{g}$ as 
\begin{equation}
    J_3(\mathbf{u},\mathbf{g}) = \lambda_g\left[\| \mathbf{u} - \mathbf{g} \|^2 + \Upsilon(\mathbf{g})\right],
\end{equation}
in which $\Upsilon(\mathbf{g})$ describes the prior distribution of $\mathbf{g}$ defined on $\mathcal{G}$. The set $\mathcal{G}$ is not convex, making it difficult to design $\Upsilon(\mathbf{g})$ explicitly. In fact, it suffcies to update $\mathbf{g}$ with fixed $\mathbf{u}$ and vise vera in the final optimization problem. With fixed $\mathbf{g}$, $\mathbf{u}$ can be easily updated with the term $\|\mathbf{u} - \mathbf{g}\|^2$. To update $\mathbf{g}$ with any fixed $\mathbf{u}\in\mathbb{R}^{I \times J \times K}$, we choose an element from the set $\mathcal{G}$ which not only lies in the neighborhood of $\mathbf{u}$ in the $L_2$ sense, but also acts like the surface of some real-world object. With this motivation in mind, we construct a map $\mathcal{S}: \mathbb{R}^{I \times J \times K} \rightarrow \mathcal{G}$ and view $\mathcal{S}(\mathbf{u})$ as the solution to the following optimization problem.
\begin{equation}
    \mathcal{S}(\mathbf{u}) = \mathop{\arg\min}_\mathbf{g} \| \mathbf{u} - \mathbf{g} \|^2 + \Upsilon(\mathbf{g}).
\end{equation}
We call $\mathcal{S}(\mathbf{u})$ a \emph{surfaciation} of $\mathbf{u}$. To construct the map $\mathcal{S}$, we assign for each $\mathbf{u} \in \mathbb{R}^{I \times J \times K}$ an indicator matrix $\mathbf{e}$, a depth matrix $\mathbf{d}$ and an albedo matrix $\bm{\alpha}$ (recall the definition of $\mathcal{G}'$). We determine these three matrices using the following approach.

For each pixel $(i,j)$ of $\mathbf{u}$, let $u_{i,j,{k^1_{ij}}},\dots,u_{i,j,{k^{n_{ij}}_{ij}}}$ be all $n_{ij}$ non-zero albedo values in the depth direction. Define
\begin{equation}
    \tilde{e}_{ij}=\left\{ \begin{array}{l}
	1,\ n_{ij} > 0\\
	0,\ n_{ij} = 0
\end{array} \right.
\end{equation}
to be the indicator function of the set of foreground pixels. There could be many mislabeled pixels due to heavy background noise in $\mathbf{u}$. To provide a noise-robust estimation of the indicator matrix $\mathbf{e}$, we consider the global correlation of all pixels and solve
\begin{equation}
    \begin{aligned}
\mathbf{e}^* = (e_{ij}^*)_{I \times J} &= \mathop{\arg\min}_{\{e_{ij}\}} \sum^I_{i=1} \sum^J_{j=1} \gamma_{ij}(e_{ij}-\tilde{e}_{ij})^2\\
 +& \sum^I_{p=1}\sum^J_{q=1} \sum^I_{r=1}\sum^J_{s=1} w^\mathbf{e}_{pq,rs}(e_{pq}-e_{rs})^2,
    \end{aligned}
\end{equation}
in which $w^\mathbf{e}_{pq,rs}$ describes the weight of the pixels $(p,q)$ and $(r,s)$. The parameter $\gamma_{ij}$ describes the confidence of indication of the originial pixel $\tilde{e}_{ij}$. This least squares problem has a unique solution and can be solved using the standard LSQR method. The indicator function $\mathbf{e} = (e_{ij})$ is determined by
\begin{equation}
    e_{ij}=\left\{ \begin{array}{l}
	1,\ e^*_{ij} \geq 0.5\\
	0,\ e^*_{ij} < 0.5
\end{array} \right..
\end{equation}

\begin{algorithm*}[htbp]
    \caption{The SSCR algorithm}
    \label{alg:SSCR}
    \begin{algorithmic}
    \REQUIRE $\mathbf{d}$, $N$.
    \ENSURE $\mathbf{u},\mathbf{g}$.
    \STATE{$\tau_{p,q}^0 = d_{p,q} / N$}
    \STATE{$\mathbf{u}^0 = \mathop{\arg\min}_{\mathbf{u}}\lambda\|\bm{\tau}^0 - A\mathbf{u}\|^2 + s_u\|\mathbf{u}\|_1$}
    \FOR{$k = 1$ to $K$}
    \STATE{$\mathbf{g}^{k+1} = \mathop{\arg\min}_{\mathbf{g}} \| \mathbf{u}^k - \mathbf{g} \|^2 + \Upsilon(\mathbf{g})$}
    \STATE{$(D^{k+1}_s,\mathbf{C}^{k+1},D^{k+1}_n) = \mathop{\arg\min}_{D_s,\mathbf{C},D_n}\|\mathcal{B}\mathbf{u}^k_i-D_sC_iD_n^T\|^2 + \lambda_{pu}|C_i|_0$}
    \STATE{$S^{k+1} = \mathop{\arg\min}_{S} \lambda_t \|\mathcal{P(\bm{\tau}}^k)- DS\|^2 + \lambda_{ut} \|\mathcal{P}(A\mathbf{u}^k)- DS\|^2 + \lambda_{pt}|S|_0$}
    \STATE{$\bm{\tau}^{k+1} = \mathop{\arg\min}_{\bm{\tau}}\sum_{p,q}\left[(d_{p,q}-N)\ln(1-\tau_{p,q}) - d_{p,q}\ln(\tau_{p,q})\right]+\lambda_t\|\mathcal{P(\bm{\tau}})- DS^{k+1}\|^2 + \lambda\|\bm{\tau} - A\mathbf{u}^k\|^2$}
    \STATE{$\mathbf{u}^{k+1} = \mathop{\arg\min}_{\mathbf{u}}\lambda\|\bm{\tau}^{k+1} - A\mathbf{u}\|^2 + s_u\|\mathbf{u}\|_1+\lambda_{ut} \|\mathcal{P}(A\mathbf{u})- DS^{k+1}\|^2+\lambda_u\sum_i\|\mathcal{B}\mathbf{u}_i-D^{k+1}_sC^{k+1}_i(D^{k+1}_n)^T\|^2+\lambda_g\|\mathbf{u}-\mathbf{g}^{k+1}\|^2$}
    \ENDFOR
    \end{algorithmic}
\end{algorithm*}

To obtain the depth matrix $\mathbf{d}$, we solve the following least squares problem

\begin{equation}
    \begin{aligned}
    \mathbf{d}^* =& (d_{ij}^*)_{I \times J}\\
     =& \mathop{\arg\min}_{\{d_{ij}\}} \sum^I_{i=1} \sum^J_{j=1} \sum^{n_{ij}}_{n=1} \lambda_{ijn}(d_{ij} - z_{k^n_{ij}})^2\\
     &+ \sum^I_{p=1}\sum^J_{q=1} \sum^I_{r=1}\sum^J_{s=1}w^\mathbf{d}_{pq,rs}(d_{pq}-d_{rs})^2,
    \end{aligned}
\end{equation}
in which $z_{k^n_{ij}}$ is the depth of the voxel $(i,j,k^n_{ij})$. $\lambda_{ijn}$ and $w^\mathbf{d}_{pq,rs}$ are fixed parameters that control the weight of the corresponding terms. The depth matrix $\mathbf{d}=(d_{ij})$ is then determined by 
\begin{equation}
    d_{ij}=\left\{ \begin{array}{l}
    \mathop{\arg\min}_{k} \|z_k - d^*_{ij}\|,\ e_{ij} = 1\\
	\text{NaN},\ \ \ \ \ \ \ \ \ \ \ \ \ \ \ \ \ \ \ \ \ \ \ \ \ \ \ e_{ij} = 0
\end{array} \right..
\end{equation}
The albedo matrix $\bm{\alpha} = (\alpha_{ij})$ is obtained by solving the following optimization problem
\begin{equation}
    \begin{aligned}
    \mathop{\min}_{\{\alpha_{ij}\}} &\sum^I_{i=1} \sum^J_{j=1} \sum^{n_{ij}}_{n=1} \lambda_{ijn}(\alpha_{ij} - u_{ijk^n_{ij}})^2\\
     &\ \ \ \ \ \ \ \ \ + \sum^I_{p=1}\sum^J_{q=1} \sum^I_{r=1}\sum^J_{s=1}w^{\bm{\alpha}}_{pq,rs}(\alpha_{pq}-\alpha_{rs})^2,
    \end{aligned}
\end{equation}
in which $\lambda_{ijn}$ and $w^{\bm{\alpha}}_{pq,rs}$ are fixed parameters. Finally, the element $\alpha_{ij}$ is reset as $\operatorname{NaN}$ if $e_{ij} = 0$.

\section{The SSCR reconstruction algorithm}
Finally we obtain the optimization problem of the proposed signal-surface collaborative regularization (SSCR) framework as follows
\begin{equation}
    \label{SSCR}
    \begin{aligned}
    &\mathop{\arg\min}_{\substack{\bm{\tau},\mathbf{u},\mathbf{g},\\D_s,D_n,\mathbf{C},S}}\sum_{p,q}\left[(d_{p,q}-N)\ln(1-\tau_{p,q}) - d_{p,q}\ln(\tau_{p,q})\right]\\
    &+ \lambda_t \|\mathcal{P}(\bm{\tau}) - DS\|^2 +  \lambda_{ut}\|\mathcal{P}(A\mathbf{u})- DS\|^2 + \lambda_{pt}|S|_0\\
    &+ \lambda\|\bm{\tau} - A\mathbf{u}\|^2 + s_u\|\mathbf{u}\|_1+ \lambda_g\left[\| \mathbf{u} - \mathbf{g} \|^2 + \Upsilon(\mathbf{g})\right]\\
    &+ \lambda_u\sum_i\left[\|\mathcal{B}\mathbf{u}_i-D_sC_iD_n^T\|^2 + \lambda_{pu}|C_i|_0\right]\\
    & s.t. \ \ g\in{\mathcal{G}},\ \ D_sD_s^T = I_x,\ \ D_nD_n^T = I_y.
    \end{aligned}
\end{equation}
in which $x$ is the product of the block size in three directions, $y$ is the number of neighbors for each block. At first sight, this problem contains many terms and seems complicated. In fact, it can be solved using the standard alternating iteration method and solutions to most sub-problems have already been well studied\cite{SOCR, CC-SOCR}. We summarize the main step of the proposed SSCR algorithm in \cref{alg:SSCR}. In the supplementary material, we provide a detailed discussion of the solution to this optimization problem and the choices of parameters.

\section{Results}

To validate the capability of the proposed method in reconstructing the hidden targets with sparse measurements, we compare our reconstruction results with the Laplacian of Gaussian filtered back-projection (LOG-BP)\cite{LOG_BP}, SOCR\cite{SOCR} and  CC-SOCR\cite{CC-SOCR} methods. We also bring the F-K \cite{FK}, LCT \cite{LCT}, and the D-LCT \cite{DLCT} methods into comparisions by constructing dense measurements with the commonly used zero-padding technique. In the supplement, we provide a gallery of reconstruction results of these methods with other signal interpolation techniques, where similar reconstruction results are shown.

\vspace{0.15cm}

\begin{figure}[t]
    \centering
     \includegraphics[width=1\linewidth]{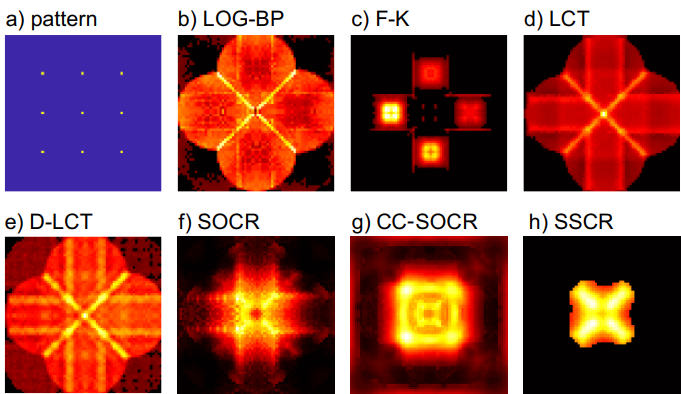}
     \caption{Comparisions of the reconstruction results of the pyramid with confocal synthetic signal. a) The signal is measured at only $3\ \times\ 3$ focal points. b) - h) The front view of the reconstructions obtained with the LOG-BP, F-K, LCT, D-LCT, SOCR, CC-SOCR, and the proposed algorithms}
     \label{fig:compare_pyramid}
  \end{figure}

\noindent\textbf{Synthetic data}\hspace{1em}We use the synthetic signal of the instance of the pyramid \cite{SOCR} to test the proposed method. Only $3 \times 3$ of the original $64 \times 64$ synthetic signals are chosen. The physical model used to generate the data considers consine attenuation of the photon intensity, and is finer than \cref{confocal_physical_model}. The base length and height of the pyramid are 1 m and 0.2 m, respectively. The central axis of the regular quadrangular pyramid is vertical to the planer relay surface. The pyramid is 0.5 m away from the relay surface and the time resolution is 32 ps. The experimental setup and the three views of our reconstruction are shown in \cref{fig:layout}. Figure \ref{fig:compare_pyramid} shows the reconstruction results of different methods. The LOG-BP, F-K, LCT and D-LCT methods fail to locate the hidden object. The SOCR and CC-SOCR methods locate the target correctly, but contain artifacts in the background. The proposed SSCR method provides a faithful estimation of the hidden target. More detailed comparisons are provided in the supplement.

  \begin{figure}[t]
    \centering
\includegraphics[width=1\linewidth]{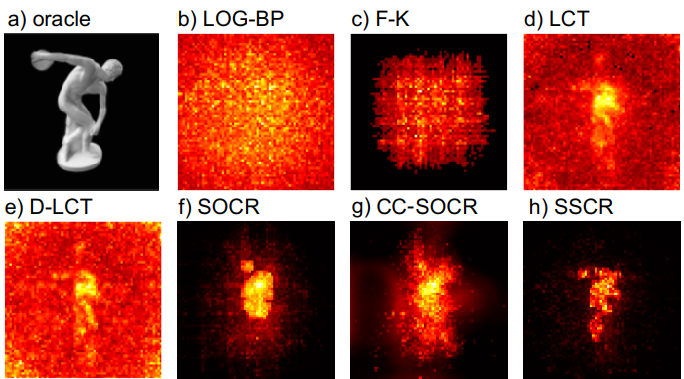}
  \caption{Comparisions of the reconstruction results of the statue with $5 \times 5$ confocal measurements\cite{FK}. a) A photo of the hidden object. b) - h) The front view of the reconstructions obtained with the LOG-BP, F-K, LCT, D-LCT, SOCR, CC-SOCR, and the proposed algorithms}
  \label{fig:compare_statue}
\end{figure}

\begin{figure}[t]
    \centering
     \includegraphics[width=1\linewidth]{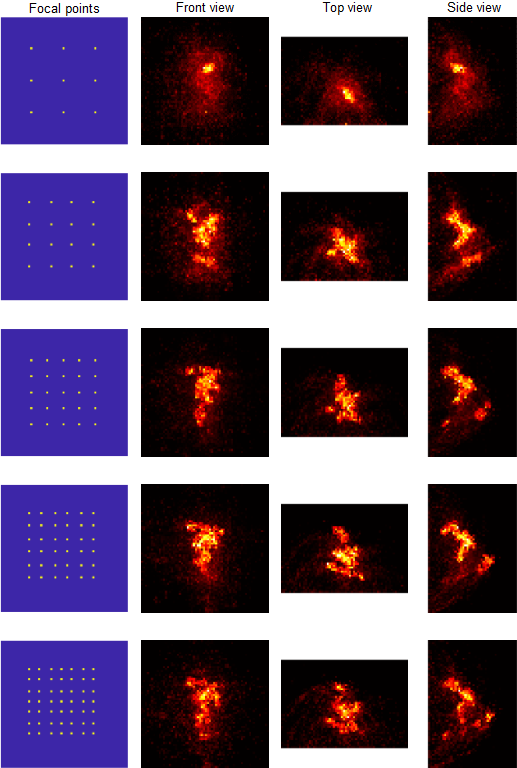}
     \caption{Reconstruction results of the proposed method with different illumination settings. The illumination points are shown in yellow in the first column. The front view, top view and side view of the reconstructions are shown in the second, third and fourth columns.}
     \label{fig6:multirec}
  \end{figure}

  \vspace{0.15cm}

  \noindent\textbf{Measured data}\hspace{1em}We use the measured data of the instance of the statue \cite{FK} to test the proposed method in real-world applications. The original dataset contains $512 \times 512$ confocal signals measured over a square region of $2 \times 2$ $\text{m}^2$ in 10 mins. The hidden target is 1 m from the visible surface. The time resolution is 32 ps. Reconstruction results with $5 \times 5$ confocal measurements are compared in \cref{fig:compare_statue}. The focal points are shown in the third row and first column of \cref{fig6:multirec}. It is shown that the LOG-BP and F-K methods fail to reconstruct the target. The LCT and D-LCT reconstructions contain heavy noise and the target can hardly be identified. The SOCR method only reconstructs a portion of the target. The CC-SOCR reconstruction is blurry and contains background noise. In contrast, the proposed SSCR method reconstructs the target faithfully. In \cref{fig6:multirec} we show three views of the SSCR reconstructions with different number of illumination points. The target can be clearly reconstructed with $7\times 7$ confocal measurements, which is only 0.01\% of the original dataset. With $3 \times 3$ illuminations, the SSCR method still provides a reasonable estimation of the hidden target, which demonstrates the robustness of the proposed algorithm. More comparisions with existing methods are provided in the supplement.

  \begin{figure}[t]
    \centering
     \includegraphics[width=1\linewidth]{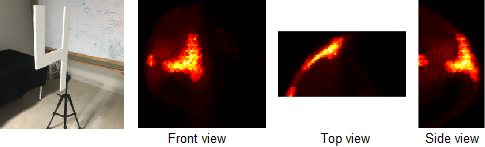}
     \caption{Reconstruction results of the figure `4' with non-confocal measurements\cite{Phasor}. The SSCR method provides an estimation of the target with only $6 \times 6$ measurements.}
     \label{fig:non_confocal}
  \end{figure}

  \vspace{0.15cm}

  \noindent\textbf{Measured data}\hspace{1em}We use the measured data for the instance of the figure `4' provided by the work \cite{Phasor} to test the SSCR method under the non-confocal setting. The target is 1 m away from the visible surface and the time resolution is 16 ps. The illumination pattern is shown in the fourth role of \cref{fig6:multirec} and the three views of our reconstruction are shown in \cref{fig:non_confocal}. Comparisions with the LOG-BP\cite{LOG_BP}, phasor field\cite{Phasor} and other methods\cite{SOCR,CC-SOCR} are provided in the supplement, where the proposed method performs the best.

\section{Discussion}

When the signal is detected at $P$ measurement pairs and the reconstruction domain is discretized with $L \times L \times L$ voxels, the time and memory complexities of the proposed SSCR method are $\mathcal{O}(PL^3)$ and $\mathcal{O}(L^3)$ respectively. In the supplement, we provide a detailed analysis of these complexities. Notebly, when $P = \mathcal{O}(1)$, the time complexity of our method is $\mathcal{O}(L^3)$. This is smaller than the F-K, LCT, and D-LCT methods, which cost $\mathcal{O}(L^3\log L)$. In this case, the SOCR method also has a time complexity of $\mathcal{O}(L^3)$ but provides biased estimations of the targets. This fact indicates the significance of the proposed two-dimensional regularization of the hidden surface. As is shown in \cref{fig:compare_pyramid} and \cref{fig:compare_statue}, the CC-SOCR method performs the second. This method relys on the dense virtual confocal signal measured over a sqaure region, which leads to a high cost of $\mathcal{O}(L^5)$. 

We also discuss some potential directions of research that help improve the proposed method. In \cref{e_tau}, we did not consider the dark count of the detector and the background noise. Modeling the noise distribution within the Bayesian framework would lead to better reconstruction results. Besides, for the two-dimensional regularization of $\mathbf{g}$,  we simply use pixel-wise similarity in the SSCR method. The two-dimensional image priors have been widely studied in the past decades, including non-local self-similarity\cite{BM3D}, low-rankness\cite{HOSVD}, low-dimensionality\cite{LDMM}, just to name a few. Incorporating these regularizations would lead to better estimations of the hidden surface. In recent years, it is shown that the family of tensor network proves to be a powerful tool for dimensionality reduction and data processing\cite{tensordata}. It would also be an interesting direction of research to reduce the dimension of the object and the signal using tensor networks.

\section{Conclusion}
We conclude that the two-dimensional surface regularization plays an important role in few-shot NLOS imaging tasks. The joint regularization of the estimated signal, the voxel-based representation of the target and the hidden surface makes it possible to reconstruct the hidden scene with certain complex geometric structures with only $5 \times 5$ confocal measurements, even in cases with high measurement noise. We would like to incorporate other powerful two-dimensional regularizations of the surface for higher noise-robustness of the SSCR framework in the future.

\vspace{0.15cm}

\noindent\textbf{Acknowledgments}\hspace{1em}This work was supported by the National Natural Science Foundation of China (61975087, 12071244, 11971258).

{\small
\bibliographystyle{ieee_fullname}
\bibliography{egbib}
}

\end{document}